\documentclass{article}
\usepackage{style_files/ijcai22}
\usepackage{bm}
\usepackage{times}

\usepackage{soul}
\usepackage{url}
\usepackage[hidelinks]{hyperref}
\usepackage[utf8]{inputenc}
\usepackage[small]{caption}
\usepackage{graphicx}
\usepackage{amsmath}
\usepackage{booktabs}
\usepackage{refcount}
\usepackage{amsfonts}
\usepackage{subfig}
\usepackage{caption}
\usepackage{color, colortbl}
\definecolor{Gray}{gray}{0.80}
\definecolor{LightGray}{gray}{0.93}

\newtheorem{pro-stat}{Problem Definition}
\newcommand{\symgnn}{\textsc{SyMGNN}}

\newcommand{\hh}[1]{{\small\color{red}{\bf HH: #1}}}
\newcommand{\bx}[1]{{\small\color{blue}{\bf BX: #1}}}
\newcommand{\hide}[1]{}
\newcommand{\mkclean}{
	\renewcommand{\hh}[1]{}
	\renewcommand{\bx}[1]{}
}
\newtheorem{problem}{Problem}

\def\T{{\scriptscriptstyle\mathsf{T}}}

\title{ Geometric Matrix Completion via Sylvester Multi-Graph Neural Network}

\author{
    Boxin Du*, Changhe Yuan$\dagger$, Fei Wang$\dagger$, Hanghang Tong*
    \affiliations
    *Department of Computer Science, University of Illinois Urbana-Champaign\\
    $\dagger$ Amazon \\
    *\{boxindu2, htong\}@illinois.edu 
    $\dagger$\{ychanghe, feiww\}@amazon.com 
}

\begin{document}
\setlength{\abovedisplayskip}{1.4pt}
\setlength{\belowdisplayskip}{1.4pt}
\setlength{\abovedisplayshortskip}{1.4pt}
\setlength{\belowdisplayshortskip}{1.4pt}
\maketitle

\mkclean

\begin{abstract}
Despite the success of the Sylvester equation empowered methods on various graph mining applications, such as semi-supervised label learning and network alignment, there also exists several limitations. The Sylvester equation's inability of modeling non-linear relations and the inflexibility of tuning towards different tasks restrict its performance. In this paper, we propose an end-to-end neural framework, \symgnn\, which consists of a multi-network neural aggregation module and a prior multi-network association incorporation learning module. The proposed framework inherits the key ideas of the Sylvester equation, and meanwhile generalizes it to overcome aforementioned limitations. Empirical evaluations on real-world datasets show that the instantiations of \symgnn\ overall outperform the baselines in geometric matrix completion task, and its low-rank instantiation could further reduce the memory consumption by $16.98\%$ on average. 
\end{abstract}

\vspace{1.0\baselineskip}
\section{Introduction}
\hide{
\bx{The outline of the intro:
Sylvester equation has shown its effectiveness in many multi-network mining tasks, such as network alignment, social recommendation, geometric matrix completion, and subgraph matching, etc. However, there are some limitations which restrict its applicability. Fist, second, third, ... . In this paper, we propose a neural framework, as an extension of the traditional linear Sylvester equation, in order to strengthen its performance in downstream tasks. Specifically we will focus on geometric matrix completion. The distinctive advantages of the proposed framework compared with both the traditional Sylvester equation and the existing neural models on geometric matrix completion are as follows.
\begin{itemize}
    \item The proposed framework is a general form, and also flexible to be instantiated towards different downstream tasks. 
    \item For geometric matrix completion, the proposed model considers the within network attention and cross-network attention, which contribute to learning compatible user and item representations.
    \item In order to further improve the spatial complexity, the low-rank model represents the completed matrix as low-rank matrices. 
\end{itemize}}
}

The Sylvester equation plays a central role for various applications in applied mathematics \cite{golub1979hessenberg} \cite{wachspress1988iterative}, systems and control theory \cite{benner2004factorized}, machine learning \cite{agovic2011probabilistic} and graph mining \cite{li2021scalable}. Particularly in graph mining, the Sylvester equation has shown its applicability in numerous multi-network mining tasks, such as network alignment \cite{zhang2016final}, social recommendation \cite{du2021sylvester}, and semi-supervised label learning \cite{chen2008semi}. 

Despite its succinct mathematical formulation, elegant theoretical properties, and numerous efficient solvers, there are several limitations when Sylvester equation is applied on multi-network mining. First, the real-world network data contains various heterogeneous features. However, it is non-trivial to directly incorporate these features of the networks into the classic Sylvester equation formulation. Second, in the task of multi-network association, the classic Sylvester equation essentially calculates a linear transformation from the observed prior multi-network association matrix. However, the non-linear relation between the prior knowledge and the final solution can not be captured by the classic Sylvester equation. Third, the Sylvester equation solver is often separated from the downstream task learning in many graph mining problems, and thus the solution of the Sylvester equation has to be further adapted towards different multi-network mining tasks. For example, in network alignment, the solution matrix of multi-network association is first calculated by the equation. Then, the soft/hard alignment method is conducted as an extra post-processing step, such as greedy match \cite{zhang2016final}. The equation can not be trained or tuned in an end-to-end fashion as modern neural networks, and consequently the performance of the downstream tasks might be suboptimal. A natural question is: {\em How can we get the best of both the traditional Sylvester equation formulation and the neural network models?}

In this paper, we propose a multi-graph neural network framework, \symgnn\, in order to generalize the traditional linear Sylvester equation towards an end-to-end neural network model. Specifically, we focus on geometric matrix completion task, and elucidate two instantiations for the \symgnn\ framework. Our proposed approach bears three distinctive advantages compared with both the Sylvester equation and the existing neural models targeted on geometric matrix completion. First, the proposed framework is a general form, and it is able to incorporate network features and be flexible to be instantiated towards different downstream tasks. Second, the neural design of the model leverages the attention mechanism, so that the proposed model incorporates both within-  and cross-network attention. This in turn helps increase the model expressiveness, capture the non-linear relations between input features, and learn compatible node representations across different networks. Third, the instantiations of the proposed framework could be trained in an end-to-end fashion, which directly adapt the solution generation module to the downstream prediction module. Fourth, for geometric matrix completion, two instantiations are provided based on explicitly learning multi-network association by 2-dimensional convolution, and learning low-dimensional representations for separate input networks, respectively. The low-dimensional instantiation approach further reduces the model's space complexity.


\section{Problem Definition}\label{sec:problem}
The notations used throughout the paper are summarized in Table \ref{tab:notations}. Generally, we use bold uppercase letters to represent matrices, bold lowercase letters to represent vectors, lowercase or uppercase letters in regular font for scalars. 
{
	\begin{table}[!thb]
		\caption{Symbols and Definition}
		\vspace{-1.0\baselineskip}
		\label{tab:notations}
		\centering
		\resizebox{0.5\textwidth}{!}{
			\begin{tabular}{|c|l|}\hline
				\textbf{Symbols} & \textbf{Definition} \\
				\hline
				$\mathcal{G}_1=\{\mathbf{A_1}, \mathbf{F_1}\}$ & a network with node feature matrix\\
				$\mathbf{H}$ & prior knowledge matrix of cross-network associations\\
				 $\mathbf{D_1}, \mathbf{D_2}$ & diagonal degree matrices\\
				$\mathbf{I}$ & an identity matrix\\
				$\mathbf{W}, \mathbf{\Theta}$ & learnable parameter matrices \\
				
				\hline
				$r$ & the dimension of node representations\\
				$d$ & the dimension of features\\
				\hline
				
				$<\mathbf{v}_i, \mathbf{v}_j>$ & the inner product of $\mathbf{v}_i, \mathbf{v}_j$ \\
				
				$\textrm{bmm}(\cdot)$ & batch matrix multiplication \\
				${\small\parallel\cdot\parallel_F}$ & Frobenius norm\\
				
				$\textrm{diag}(\mathbf{v})$ & construct a diagonal matrix by vector $\mathbf{v}$ \\
			
				\hline \end{tabular}
				}\end{table}
}

Before giving the definition of GNN-based neural Sylvester equation in Section \ref{sec:method}, we first provide some preliminaries on the traditional Sylvester equation and the Graph Neural Networks, followed by a formal definition of the geometric matrix completion.

\subsection{Preliminaries}
\noindent \textbf{A - Sylvester Equation for Multi-network Mining.} 
Given two networks represented as $\mathcal{G}_1=\{\mathbf{A}_1, \mathbf{F}_1\}$, $\mathcal{G}_2=\{\mathbf{A}_2, \mathbf{F}_2\}$, and an anchor multi-network association matrix $\mathbf{H}$, which denotes the prior knowledge of the multi-network node associations. The Sylvester equation for multi-network mining is defined as follows \cite{du2018fasten}:
\begin{equation}\label{eq:sylvester}
    \mathbf{X} = \alpha\mathbf{\Tilde{A}}_2\mathbf{X}\mathbf{\Tilde{A}}^{\T}_1 + (1-\alpha)\mathbf{H}
\end{equation}
where $\mathbf{\Tilde{A}}_1$ and $\mathbf{\Tilde{A}}_2$ are the symmetrically normalized adjacency matrices of the input networks. The $\mathbf{X}$ represents the cross-network node association scores which the equation aims to calculate. The scalar $\alpha\in(0,1)$ is aimed at weighting the multi-network association aggregation term (i.e. $\mathbf{\Tilde{A}}_2\mathbf{X}\mathbf{\Tilde{A}}^{\T}_1$), and the prior knowledge term ($\mathbf{H}$). Due to the normalization of $\mathbf{A}_1$ and $\mathbf{A}_2$, the corresponding linear system of Eq. \eqref{eq:sylvester} contains a positive semi-definite coefficient matrix, which guarantees the existence of unique solution for Eq. \eqref{eq:sylvester}. Solving Eq. \eqref{eq:sylvester} is often time-consuming. A straightforward iterative method to solve Eq. \eqref{eq:sylvester} is the fixed point iteration. More efficient method is proposed in \cite{du2018fasten} with linear time and space complexity.

The formulation of this equation for multi-network mining (Eq. \eqref{eq:sylvester}) enjoys several distinctive advantages, which are summarized as follows. Firstly, theoretically the existence and uniqueness of the solution $\mathbf{X}$ can be guaranteed. Furthermore, there exists various efficient and scalable solvers for the solution. Secondly, the solution $\mathbf{X}$ can be seen as a fixed point of the equation and can be obtained by iteratively evaluating the Eq. \eqref{eq:sylvester}. Compared to existing neural models, which might contain a number of hidden layers, there is no need to save the hidden states/representations. Thirdly, when reaching the fixed point, theoretically it is equivalent to proceed the recurrent process implied by Eq. \eqref{eq:sylvester} infinite times, so the formulation is able to leverage long-range dependency when solving $\mathbf{X}$.

However, despite the advantages and effectiveness in various tasks, generally there are also several limitations of this formulation which are summarized as follows. Firstly, the numerical features of the nodes can not be effectively utilized for calculating $\mathbf{X}$. Secondly, the $\mathbf{X}$ can be seen as a linear transformation from the prior knowledge matrix $\mathbf{H}$. However, the potential non-linear relationship between them can not be captured by this formulation. Thirdly, since the equation is not learnable and not tunable, the solution $\mathbf{X}$ should always be adapted to a target downstream task by another learning model, but not in an end-to-end fashion. This might result in suboptimal performance for the downstream task. 

\noindent \textbf{B - Graph Neural Networks.} The Graph Neural Networks (GNN) are powerful deep learning models for network data. The basic idea of GNN model is to learn node representations via learnable aggregation, in which the node features are accumulated and transformed from the neighborhood features. Given a network $\mathcal{G} = (\mathbf{A}, \mathbf{F})$, where $\mathbf{A}\in\mathbb{R}^{n\times n}$ is the adjacency matrix of $\mathcal{G}$, and $\mathbf{F}\in\mathbb{R}^{n\times d}$ is the feature matrix with $d$ being the dimension of features, representative GNN aggregation step at time step $t$ can be written as follows. 
\begin{equation}
    \mathbf{X}^{(t+1)} = \phi(\mathbf{\tilde{A}}\mathbf{X}^{(t)}\mathbf{W} + \mathbf{\Omega}^{(t)}\mathbf{F})
\end{equation}
where $\mathbf{\tilde{A}}$ is the normalized adjacency matrix with added self-loops. $\mathbf{W}$ is a learnable parameter matrix for the aggregated features. Different GNN models adopts different feature aggregation mechanisms. GCN model \cite{kipf2016semi} inserts the adjacency matrix with self-links and applies the re-normalization. It also sets $\mathbf{\Omega} = \mathbf{0}$. GAT model \cite{velivckovic2017graph} utilizes the self-attention mechanism in feature aggregation. GIN model \cite{xu2018powerful} adopts an MLP layer after the aggregation of hidden representations of nodes for improving the discriminative ability of GNN model. 

\noindent \textbf{C - Convolutional Graph Embedding.} 
Proposed in \cite{yao2018convolutional}, the convolutional graph embedding (CGE) model is a GNN-based single network embedding model. Different from traditional GCN \cite{kipf2016semi}, which simply sums up the hidden representations of all neighbors, the aggregation weights for center nodes and neighbor nodes are differentiated and learnable with the model in CGE. Specifically, in the $(l+1)$-th layer, the output of a CGE layer can be represented as:
\begin{equation}\label{eq:cge}
    \mathbf{V}^{(l+1)} = \phi((\textrm{diag}(\bm{\sigma}) + (\mathbf{I} - \textrm{diag}(\bm{\sigma}))\mathbf{\Tilde{A}})\mathbf{V}^{(l)}\mathbf{\Theta}^{(l)})
\end{equation}
where $\mathbf{V}^{(l+1)}$ and $\mathbf{V}^{(l)}$ are the node representation matrices of the $(l+1)$-th and $l$-th layer, respectively. $\bm{\sigma}$ is the learnable weight vector for the self-connections, and $\mathbf{\Theta}^{(l)}$ is the learnable weight matrix. $\phi()$ is an activation function. Using CGE adds more expressiveness to the model compared with GCN, and we will elaborate how to leverage it in Section \ref{sec:method}.

\subsection{Geometric Matrix Completion}
Different from traditional matrix completion problem, the geometric matrix completion needs to handle two additional networks which reflect the topological relations between the nodes of two entities sets. Specifically, the problem is defined as follows. 
\begin{problem}\textsc{Geometric Matrix Completion} \\\label{prob:gmc}
    \textbf{Given:} Two networks with node features $\mathcal{G}_1=\{\mathbf{A}_1, \mathbf{F}_1\}$, $\mathcal{G}_2=\{\mathbf{A}_2, \mathbf{F}_2\}$, and the partially observed multi-network association $\mathbf{H}$ of the nodes in $\mathcal{G}_1$ and $\mathcal{G}_2$; \\
    \textbf{Output:} The unobserved entries in $\mathbf{H}$.
\end{problem}

\section{Proposed Model}\label{sec:method}
In this section, we elaborate our proposed framework of multi-graph neural networks. First, we present a general framework of neural Sylvester equation, followed by two GNN-based instantiations of the general framework targeted at the geometric matrix completion. Second, we introduce the training method with details. Finally, we provide analysis of the proposed two instantiations in terms of computatinal complexity.
\begin{figure*}
    \centering
    \includegraphics[width=1.0\textwidth, height=0.43\textwidth]{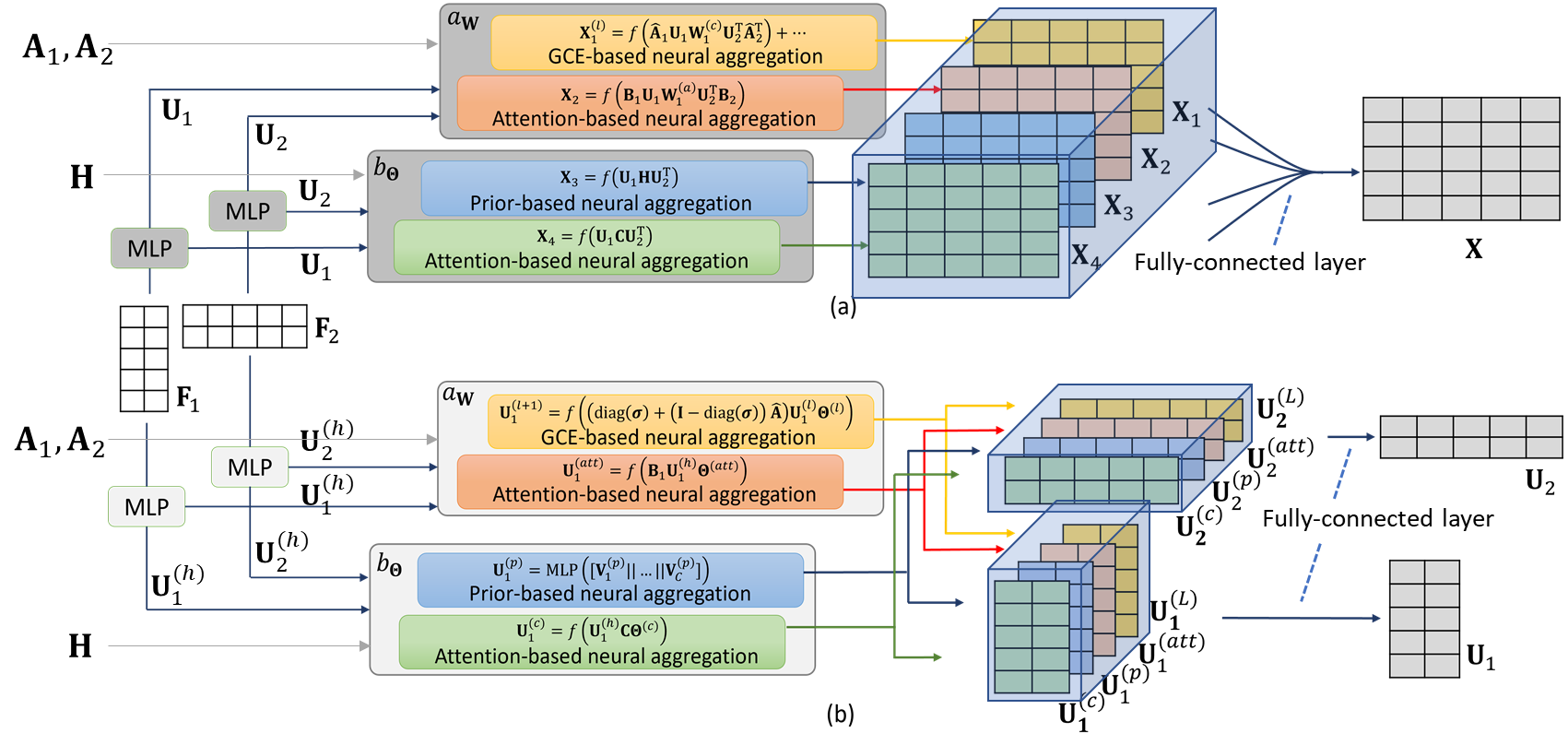}
    \vspace{-1.8\baselineskip}
    \caption{The overall illustration of two instantiations for \symgnn. (a): the base model, and (b): the low-rank model.}
    \label{fig:model}
\end{figure*}
\subsection{Sylvester Multi-Graph Neural Network Framework}
The goal of the proposed \symgnn\ framework is to leverage the advantages of the traditional Sylvester equation, and in the meanwhile overcoming its limitations. First, if we observe the Sylvester equation in Eq. \eqref{eq:sylvester} from an iterative perspective, we can see that the first term on the right side aggregate the multi-network node association $\mathbf{X}$ linearly for the updated $\mathbf{X}$. The second term incorporates the prior multi-network association message $\mathbf{H}$ into the updated $\mathbf{X}$. Second, similar to the ideas of the traditional Sylvester equation, we identify the two key modules of the \symgnn: (1) the multi-network aggregation learning module; and (2) the prior multi-network association incorporation learning module. The general framework can be represented in Eq. \eqref{eq:nse}.
\begin{equation}\label{eq:nse}
    \mathbf{X} = \phi(\alpha \cdot a_{\mathbf{W}}(\mathbf{F}_1, \mathbf{F}_2, \mathbf{\Tilde{A}}_1, \mathbf{\Tilde{A}}_2) + (1-\alpha)\cdot b_{\mathbf{\Theta}}(\mathbf{F}_1, \mathbf{F}_2, \mathbf{H}))
\end{equation}
where $a_{\mathbf{W}}()$ and $b_{\mathbf{\Theta}}()$ are two neural modules with parameters $\mathbf{W}$ and $\mathbf{\Theta}$, and weighting scalar $\alpha \in [0,1]$. $\phi()$ is a non-linear activation function. $\mathbf{X}$ is the multi-network association output of the \symgnn\ framework, and it can be further fed into a neural network for adapting towards a downstream task in an end-to-end fashion. As we can see, this framework is a neural generalization originated from the Sylvester equation in Eq. \eqref{eq:sylvester}. When the neural modules $a_{\mathbf{W}}$ and $b_{\mathbf{\Theta}}$ are linear aggregation functions, the Eq. \eqref{eq:nse} degenerates to the classic Sylvester equation Eq. \eqref{eq:sylvester}. Numerous instantiations exist for different downstream tasks. Next, we will discuss how to specifically instantiate this framework towards geometric matrix completion (Problem~\ref{prob:gmc}).

\subsection{Base Model for Geometric Matrix Completion}
Here, we present our base model for geometric matrix completion problem. In order to instantiate $a_{\mathbf{W}}(\mathbf{F}_1, \mathbf{F}_2, \mathbf{A}_1, \mathbf{A}_2)$, we design two parallel layers which adopt the CGE-based aggregation layer and the attention-based aggregation layer respectively. The motivation here is to learn the 2-d hidden representations for the multi-network association solution. In order to achieve this, we design two types of 2-d convolutional non-linear aggregation module, namely the adjacency matrix-based GCE neural aggregation, and the attention-based neural aggregation. To be specific, given $\mathcal{G}_1=\{\mathbf{A}_1, \mathbf{F}_1\}, \mathcal{G}_2=\{\mathbf{A}_2, \mathbf{F}_2\}$ with $n_1, n_2$ nodes respectively, we first apply the learnable parameters of self-connections from Eq. \eqref{eq:cge} on $\mathbf{A}_1$ and $\mathbf{A}_2$ to obtain the updated adjacency matrices. The goal is to adopt the learnable weight of the self-connections from CGE for improving the expressiveness of the model: 
\begin{subequations}
    \begin{equation}
        \mathbf{\hat{A}}_1 = \textrm{diag}(\bm{\sigma_1}) + (\mathbf{I} - \textrm{diag}(\bm{\sigma_1}))\mathbf{\Tilde{A}}_1
    \end{equation}
    \vspace{-1.0\baselineskip}
    \begin{equation}
        \mathbf{\hat{A}}_2 = \textrm{diag}(\bm{\sigma_2}) + (\mathbf{I} - \textrm{diag}(\bm{\sigma_2}))\mathbf{\Tilde{A}}_2
    \end{equation}
\end{subequations}
where $\bm{\sigma_1}$ and $\bm{\sigma_2}$ are learnable weights for self-connections of the first network and the second network respectively. 
Before applying the multi-network aggregation layers, the node features of $\mathcal{G}_1$ and $\mathcal{G}_2$ are fed into an MLP layer for obtaining the hidden features $\mathbf{U}_1=\textrm{MLP}_1(\mathbf{F}_1)$ and $\mathbf{U}_2=\textrm{MLP}_2(\mathbf{F}_2)$.
In the CGE-based multi-network aggregation, the output of the $l$-th level aggregation can be represented as: 
\begin{equation}
    \mathbf{X}_1^{(l)} = \sum_{i=1}^{l} \phi(\mathbf{\hat{A}}^{i}_1\mathbf{U}_1\mathbf{W}^{(c)}_i\mathbf{U}_2^{\T}(\mathbf{\hat{A}}^{i}_2)^{\T})
\end{equation}
where $\mathbf{W}^{(c)}_1, \cdots \mathbf{W}^{(c)}_l$ are parameter matrices and $\phi()$ is an activation function. In practice, $\mathbf{W}^{(c)}_i, i=1,2,...,l$ is implemented as $\mathbf{W}^{(c)}_i = \mathbf{W}'_i(\mathbf{W}'_i)^{\T}$, as a metric learning approach for the generalization of Mahalanobis distance \cite{yoshida2021distance}, in order to capture the feature correlation between nodes from two different networks. 

In the attention-based multi-network aggregation, the output can be represented as:
\begin{equation}
    \mathbf{X}_2 = \phi(\mathbf{B}_1\mathbf{U}_1\mathbf{W}^{(a)}\mathbf{U}_2^{\T}\mathbf{B}_2^{\T})
\end{equation}
where $\mathbf{W}^{(a)}$ is the parameter matrix, $\mathbf{B}_1$ and $\mathbf{B}_2$ are attention score matrices for $\mathcal{G}_1$ and $\mathcal{G}_2$ respectively. For instance, the attention score of node $(i,j)\in \mathcal{G}_1$ is calculated as:
\begin{equation}\label{eq:attn}
    \mathbf{B}_1(i,j) = \frac{\textrm{exp}(<\mathbf{u}_i, \mathbf{u}_j>)}{\sum_{k=1}^{n_1}\textrm{exp}(<\mathbf{u}_i,\mathbf{u}_k>)}
\end{equation}

In order to instantiate $b_{\mathbf{\Theta}}(\mathbf{F}_1, \mathbf{F}_2, \mathbf{H}))$, similar to the first term $a_{\mathbf{W}}(\mathbf{F}_1, \mathbf{F}_2, \mathbf{\Tilde{A}}_1, \mathbf{\Tilde{A}}_2)$, we can also adopt two types of parallel aggregation layers. The first one is the direct neural aggregation from prior multi-network association, and the second one is via attention schema. However, since the entries of the prior multi-network association matrix $\mathbf{H}$ is often real values or multi-class categorical rates, it is unreasonable to directly use $\mathbf{H}$ for cross-network feature aggregation. Thus the prior multi-network association-based multi-network aggregation is only adopted when $\mathbf{H}$ contains binary associations. The two types of multi-network aggregation modules are shown as follows.
\begin{subequations}
    \begin{equation}
    \mathbf{X}_3 = \phi(\mathbf{U}_1\mathbf{H}\mathbf{U}_2^{\T})
\end{equation}
    \vspace{-1.0\baselineskip}
\begin{equation}\label{eq:x4}
    \mathbf{X}_4 = \phi(\mathbf{U}_1\mathbf{C}\mathbf{U}_2^{\T})
\end{equation}
\end{subequations}
where the cross-network attention score matrix $\mathbf{C}$ is calculated as $\mathbf{C}(i,j) = \frac{\textrm{exp}(<\mathbf{u}_i, \mathbf{u}_j>)}{\sum_{k=1}^{n_2}\textrm{exp}(<\mathbf{u}_i,\mathbf{u}_k>)}$ for $i\in\mathcal{G}_1, j\in\mathcal{G}_2$. 

Putting everything together, as shown in Figure \ref{fig:model}, the intermediate multi-network association matrices $\mathbf{X}_1, \mathbf{X}_2, \mathbf{X}_3, \mathbf{X}_4$ consist of the hidden representation tensor $\bm{\mathcal{X}}\in\mathbb{R}^{n_1\times n_2\times 4}$ for the multi-network association solution. We apply a fully connected layer on $\bm{\mathcal{X}}$ for obtaining the final multi-network association $\mathbf{X} = \textrm{bmm}(\bm{\mathcal{X}, \bm{\mathcal{W}}})$ where $\bm{\mathcal{W}}\in\mathbb{R}^{n_1\times 4\times1}$ is the parameter tensor.

\subsection{Low-rank Model for Geometric Matrix Completion}
Instead of conducting bi-linear neural aggregation for the multi-network association directly, we can generate the embeddings for nodes of $\mathcal{G}_1$ and $\mathcal{G}_2$ respectively. Similar to the base model, we consider both the direct neural aggregation from the original network topology, and the neural aggregation from the within network attentions. First, given two networks $\mathcal{G}_1=\{\mathbf{A}_1, \mathbf{F}_1\}, \mathcal{G}_2=\{\mathbf{A}_2, \mathbf{F}_2\}$ with $n_1, n_2$ nodes respectively, the node features are fed into an MLP layer for obtaining the hidden features $\mathbf{U}^{(h)}_1=\textrm{MLP}^{(l)}_1(\mathbf{F}_1)$ and $\mathbf{U}^{(h)}_2=\textrm{MLP}^{(l)}_2(\mathbf{F}_2)$. Similar to the motivation of the base model, $\mathbf{U}^{(h)}_1, \mathbf{U}^{(h)}_2$ are then fed into two parallel CGE-based and attention-based neural modules for generating the hidden representations of the node features for two networks separately. We take $\mathbf{U}^{(h)}_1$ as an example, and the process for $\mathbf{U}^{(h)}_2$ is similar. The updated node hidden representations after an $l$-layer CGE module is:
\begin{equation}
    \mathbf{U}^{(l+1)}_1 = \phi((\textrm{diag}(\bm{\sigma}) + (\mathbf{I} - \textrm{diag}(\bm{\sigma}))\mathbf{\Tilde{A}})\mathbf{U}^{(l)}_1\mathbf{W}^{(l)})
\end{equation}
where $\mathbf{U}_1^{(0)} = \mathbf{U}^{(h)}_1$, $\mathbf{\Theta}^{(l)}$ is the parameters for the $l$-th layer, and $\phi()$ is an activation function. After $L$ layers, we obtain $\mathbf{U}_1^{(L)}$.
The updated node hidden representations after the attention-based neural aggregation module is:
\begin{equation}
    \mathbf{U}_1^{(att)} = \phi(\mathbf{B}_1\mathbf{U}^{(h)}_1\mathbf{W}^{(att)})
\end{equation}
where the attention score matrix $\mathbf{B}_1$ can be calculated via Eq. \eqref{eq:attn}.  $b_{\mathbf{\Theta}}(\mathbf{F}_1, \mathbf{F}_2, \mathbf{H})$ is also instantiated for $\mathcal{G}_1$ and $\mathcal{G}_2$ separately. Here, we can adopt a similar prior multi-network association-based neural aggregation when the prior $\mathbf{H}$ denotes binary or multi-class relations. For $\mathbf{H}$ with entries of $K$ classes, we apply $K$ neural networks, in which each neural network aggregates one class of nodes. 
\begin{subequations}
    \begin{equation}
        \mathbf{V}^{(p)}_i = \phi(\mathbf{H}_i\mathbf{U}^{(h)}_1\mathbf{\Theta}^{(p)}_i)
    \end{equation}
    \vspace{-1.0\baselineskip}
    \begin{equation}
        \mathbf{U}_1^{(c)} = \phi(\mathbf{C}\mathbf{U}^{(h)}_1\mathbf{\Theta}^{(c)})
    \end{equation}
\end{subequations}
where $\mathbf{H}_i$ is the prior multi-network association which only contains the entries of the $i$-th class. $\mathbf{\Theta}_i^{(p)}$ and $\mathbf{\Theta}^{(c)}$ are learnable parameters. The $\mathbf{V}_i^{(p)}$ for all classes are then concatenated and fed into an MLP for the node representation $\mathbf{U}_1^{(p)}=\textrm{MLP}([\mathbf{V}_1^{(p)}||\cdots||\mathbf{V}_K^{(p)}])$. The cross-network attention matrix $\mathbf{C}$ is calculated by the same method as in Eq. \eqref{eq:x4}.

Putting everything together, we now have four representation matrices for each network: $\mathbf{U}_1^{(L)}, \mathbf{U}_1^{(att)}, \mathbf{U}_1^{(p)}, \mathbf{U}_1^{(c)}$. We can adopt another fully connected layer to obtain a final representation $\mathbf{U}_1$\footnote{We find that by simply adding them with the original node hidden representations, we can already achieve superior performance.}. The predicted multi-network association between two nodes is calculated by the dot product of the row vectors of the resulting representation matrices $\mathbf{U}_1$ and $\mathbf{U}_2$.

\subsection{Training}
For matrix completion, we adopt the Mean Squared Error (MSE) loss for both instantiations:
\begin{subequations}
    \begin{equation}
        \mathcal{L}_1 = ||\mathbf{H} - \mathbf{M}\odot\mathbf{X}||^2_F
    \end{equation}
    \vspace{-1.0\baselineskip}
    \begin{equation}
        \mathcal{L}_2 = ||\mathbf{H} - \mathbf{M}\odot (\mathbf{U}_1\mathbf{U}_2^{\T})||^2_F
    \end{equation}
\end{subequations}
where the $\mathbf{M}$ matrix is a mask of $0, 1$, with $1$ indicating the position of the observed prior multi-network associations.
For the low-rank instantiation, the dot product of the node representations are used as the final solutions. For the regularization of the model parameters, we adopt the weight decay method with 0.01 as decay factor as we find that it shows slightly better performance over $L_2$ regularization. We use the Adam optimizer as it overall shows the stablest training.

\subsection{Complexity Analysis}
For notation simplicity, assume that the two input networks contain $n$ nodes and $m$ edges respectively. Suppose the feature dimension is $d$, the number of observed rating is $m'$ and the dimension of node representations is $r<d$. For the base model, the major computation lies in the within-network and cross-network attention calculation as well as the aggregation. From Eq. \eqref{eq:attn}, the within-network attention aggregation costs $O(n^2d)$. From Eq. \eqref{eq:x4}, the cross-network attention aggregation costs $O(n^2d)$. Eq. \eqref{eq:cge} costs $O(Lmd+d^2n)$. Since usually $r, d << m, n$, so its computation is not comparable with the attention-based neural aggregation. The overall time complexity for the base model is $O(\#iter\cdot(n^2d))$, where $\#iter$ is the total number of iterations. The space complexity is $O(n^2)$ because of the main storage of attention score matrices. Similarly, for the low-rank model, the overall time and space complexity are also $O(\#iter\cdot(n^2d))$ and $O(n^2)$. However, for the low-rank model, if we do not apply the within-network and cross-network attention-based neural aggregation, the time and space complexity would be reduced to $O(L(md+nd^2)+m'd)$, and $O(m + n(d^2+r^2))$ respectively. As we will see from Section \ref{sec:experiments}, this will reduce the running time with slight performance drop. Since the base model needs to store the intermediate multi-network association matrix, the space complexity can not be further reduced even if the attention-based neural aggregation is dropped.  
\section{Experiments}\label{sec:experiments}
In this section, we present the experimental results on real-world benchmark datasets to show the effectiveness of the proposed models. 

\subsection{Experimental Setting}
\noindent \textbf{A - Datasets and Pre-processing.} The benchmark datasets used in the experiments are summarized in Table \ref{tab:dataset}. 
\vspace{-0.5\baselineskip}
\begin{table}[h]
    \centering
    \caption{The statistics of the benchmark datasets.}
    \vspace{-1.0\baselineskip}
    \label{tab:dataset}
    \fontsize{8.8}{11}\selectfont
    \begin{tabular}{c|c|c|c|c}
    \hline
         Dataset & \# of Users & \# of Items & \# of Ratings & Density  \\
         \hline
         Douban & 3,000 & 3,000 & 136,891 & 0.0152 \\
         \hline
         Flixster & 3,000 & 3,000 & 26,173 & 0.0029 \\
         \hline
         YahooMusic & 3,000 & 3,000 & 5,335 & 0.0006 \\
         \hline
        
         ML-100K & 943 & 1,682 & 100,000 & 0.0630 \\
         \hline
         ML-1M & 69,878 & 10,677 & 1,000,209 & 0.0447 \\
         \hline
    \end{tabular}
\end{table}
\vspace{-0.5\baselineskip}

For the benchmark datasets, ML-3K and Flixster have both user-user and item-item interaction networks. Douban only contains a user-user interaction network and YahooMusic only contains an item-item interaction network. For these two datasets, we use the identity matrix as the adjacency matrix for the missing networks. For ML-100K, ML-1M, we construct their user-user and item-item interaction networks by adopting a {\em k}-nearest neighbors search via their features, and {\em k} is treated as a hyperparameter in our model. All the datasets include multi-class categorical ratings. For the training/testing split, we use the same partition which is also adopted by existing methods, such as \cite{yao2018convolutional} \cite{monti2017geometric}, etc.

\noindent \textbf{B - Baseline Methods.}
We use five baselines in our comparison, including the traditional Sylvester equation \textit{Sylv.} \cite{wachspress1988iterative}, and recent neural network-based and GNN-based methods: \textit{IGMC} \cite{zhang2019inductive}, GC-MC \cite{berg2017graph}, PinSage \cite{ying2018graph}, and sRGCNN \cite{monti2017geometric}.

\noindent \textbf{C - Experimental and Hyperparameter Settings.}
For the effectiveness comparison, we tune the hyperparameters of the model based on the best performance on the validation set. We use 2-layer GCE and attention aggregation in both instantiations on all datasets except for ML-100K and ML-1M. On these two datasets, the base model uses 3-layer GCE and attention aggregation. For the {\em k}-NN method used for generating social networks and item-item interaction networks on ML-100K and ML-1M datasets, we use $k=10$ for the low-rank model and $k=12$ for the base model. Further studies of the sensitivity of $k$ will be discussed in the ablation study. The metric for comparison is the widely adopted rooted mean squared error (RMSE).

\subsection{Effectiveness Results}
The first comparison results are shown in Table \ref{tab:comparison}, as these datasets are the most common benchmarks among all existing methods. The results are reported based on the average of five runs. The best performances are shown in bold fonts and the second best performances are shown with underlines. As we can observe from the table, the traditional Sylvester equation can not achieve competitive results compared to other neural network/GNN-based baseline methods, which is consistent with our discussion on the limitations of the Sylvester equation. The Sylvester equation can not effectively incorporate node features, and also can not capture non-linear relations between the observed multi-network association and the solution. Among all the neural network-based methods, the proposed framework with low-rank instantiation outperforms the rest of the baselines on Douban, Flixster and YahooMusic datasets. Flixster (U) represents the dataset with only the usage of user-user interaction network. The performance of the proposed method slightly drops, and it shows the importance of both interaction networks of users and items in our model. Particularly, on YahooMusic dataset, the proposed method achieves $7.65\%$ improvement over the best baseline. The average improvement over all datasets on Douban, Flixster and YahooMusic is $2.58\%$, which shows the effectiveness of the proposed models. 
\vspace{-0.5\baselineskip}
\begin{table}[h]
    \centering
    \caption{RMSE comparison for geometric matrix completion.}
    \vspace{-1.0\baselineskip}
    \label{tab:comparison}
    \fontsize{9}{11}\selectfont
    \begin{tabular}{c|c|c|c|c}
    \hline
        Method & Douban & Flixster & Flixster (U) & Yahoo \\
        \hline
         Sylv. & 1.220 & 1.244 & 1.276 & 29.403   \\
         \hline
         IGMC & \underline{0.729} & \underline{0.895} & \textbf{0.895} & 19.292 \\
         \hline
         GC-MC & 0.734 & 0.917 & 0.941 & 20.501  \\
         \hline
         PinSage & 0.739 & 0.954 & 0.951 & 22.954  \\
         \hline
         sRGCNN & 0.801 & 0.926 & 1.179 & 22.415  \\
         \hline
         Ours (base) & 0.762 & 0.911 & 0.934 & \underline{19.277}  \\
         \hline
         Ours (low-rank) & \textbf{0.725} & \textbf{0.891} & \underline{0.916} & \textbf{17.815}  \\
         \hline
    \end{tabular}
    
\end{table}

    

The comparison results on ML-100K and ML-1M datasets are shown in Table \ref{tab:ml-100k}. 
As we can see, the Sylvester equation is still not competitive with the rest of the methods. Our proposed low-rank instantiation consistently performs the best over all baselines. Among all baselines, \textit{GC-MC} has close performance compared with our methods. \textit{GC-MC} contains the graph encoder and the bi-linear decoder architecture which has similar effects as our proposed GNN-based neural aggregation model. This is consistent with our intuition of the effectiveness of the cross-network feature aggregation. 
\vspace{-0.5\baselineskip}
\begin{table}[h]
    \centering
     \caption{RMSE comparison on ML-100K and ML-1M dataset.}
     \vspace{-1.0\baselineskip}
    \label{tab:ml-100k}
    \begin{tabular}{c|c|c}
    \hline
        Method & ML-100K & ML-1M \\
        \hline
         Sylv. & 1.403 & 1.323 \\
         \hline
         IGMC & 0.922 & 0.857 \\
         \hline
         GC-MC & \underline{0.905} & 0.854 \\
         \hline
         PinSage & 0.942 & 0.906 \\
         \hline
         sRGCNN & 0.931 & 0.865 \\
         \hline
         Ours (base) & 0.915 & \underline{0.851} \\
         \hline
         Ours (low-rank) & \textbf{0.899} & \textbf{0.843} \\
         \hline
    \end{tabular}
\end{table}
\vspace{-1.0\baselineskip}

\subsection{Ablation Study}
The ablation study results are shown in Table \ref{tab:ablation}. The `Base model (G)' and `Base model (A)' represent the model with only GCE-based neural aggregation and the model with only attention-based neural aggregation. The low-rank model uses the same abbreviation. The values inside the parentheses denote the maximum allocated GPU memory in one epoch, in which we use the same batch size (i.e. 50) for comparison. As we can see, firstly the original model performs the best over all variants in terms of RMSE for both base and low-rank instantiations. Secondly, the model without the attention neural aggregation overall consumes the least GPU memory during training. On average, with only $1.24\%$ performance drop, the models without attention neural aggregation show $16.98\%$ less memory consumption. Furthermore, comparing with other baselines' performance in Table \ref{tab:ml-100k}, the variant low-rank model in Table \ref{tab:ablation} still outperforms all baseline methods. 
\vspace{-1.0\baselineskip}
\begin{table}[h]
    \centering
     \caption{Ablation study on ML-100K and ML-1M dataset.}
     \vspace{-1.0\baselineskip}
    \label{tab:ablation}
    \begin{tabular}{c|c|c}
    \hline
        Method & ML-100K & ML-1M \\
        \hline
         Base model & \textbf{0.915} (160Mb) & \textbf{0.851} (2,371Mb) \\
         \hline
         Base model (G) & 0.932 (141Mb) & 0.862 (2,099Mb) \\
         \hline
         Base model (A) & 0.924 (148Mb) & 0.861 (2,209Mb) \\
         \hline
         \hline
         Low-rank & \textbf{0.899} (136Mb) & \textbf{0.843} (1,983Mb) \\
         \hline
         Low-rank (G) & 0.902 (110Mb) & 0.857 (1,476Mb) \\
         \hline
         Low-rank (A) & 0.920 (116Mb) & 0.853 (1,641Mb) \\
         \hline
    \end{tabular}
\end{table}
\vspace{-1.0\baselineskip}

\subsection{Parameter Sensitivity}

\begin{figure}[h]
\hspace*{\fill}%
\begin{minipage}[h]{0.23\textwidth}
\centering
\vspace{0pt}
\includegraphics[width=1\textwidth, height=0.8\textwidth]{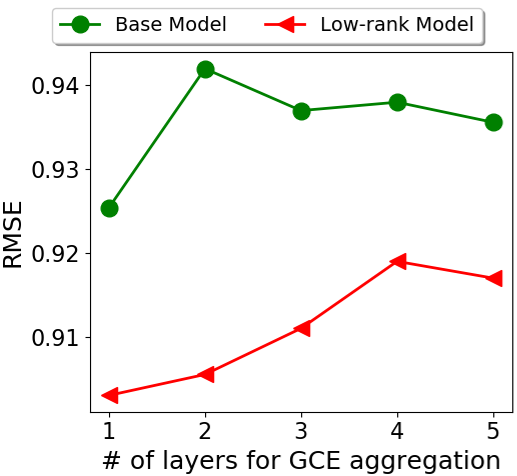}
\vspace{-1.7\baselineskip}
\caption{RMSE vs. number of layers for GCE aggregation.}
\label{fig:gnn_parameter}
\end{minipage}%
\hspace{2.00mm}
\begin{minipage}[h]{0.23\textwidth}
\centering
\vspace{0pt}
\includegraphics[width=1\textwidth, height=0.8\textwidth]{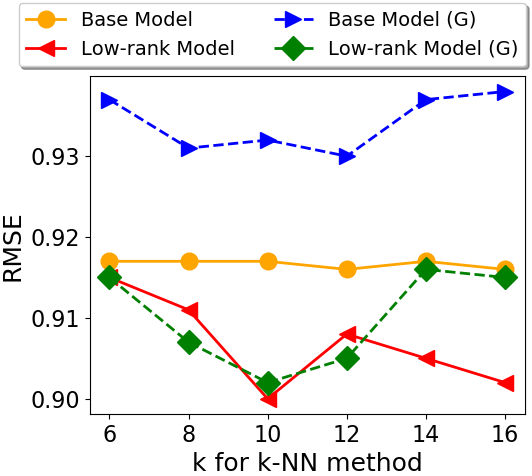}
\vspace{-1.7\baselineskip}
\caption{RMSE vs. k for k-NN method in graph construction.}
\label{fig:knn_parameter}
\end{minipage}%
\vspace{-1.0\baselineskip}
\end{figure}

We mainly study the impact of the number of GCE-based neural aggregation layers and the $k$ value for {\em k}-NN method in graph construction in ML-100K datasets. The results are shown in Figures \ref{fig:gnn_parameter} and \ref{fig:knn_parameter}. From Figure \ref{fig:gnn_parameter}, we can see that the performance is relatively stable for both models in terms of different number of layers. From Figure \ref{fig:knn_parameter}, the original models exhibit stabler performance over the models without the attention-based neural aggregation (the dashed lines) w.r.t. the $k$ value. This also indicates that the attention-based neural aggregation makes the model less sensitive to the hyperparameter for constructing graphs when the user-user/item-item interactions are not directly available. 
\vspace{-1.0\baselineskip}

\section{Related work}\label{sec:related-work}

\noindent \textbf{A - Multi-network Mining.} 
Generally, multi-network mining techniques can be categorized into traditional numerical techniques and recent neural techniques. For numerical methods, \textit{GT-COPR} by Li et al. \cite{li2019learning} aims at inferring multi-relations among the entities across multiple networks by a low-rank tensor by a tensor-based optimization method. After that, Li et al. \cite{li2021scalable} propose an optimization
method and a low-rank tensor-based label propagation algorithm for multi-relation inference. Liu et al. propose a cross-network multi-relation association learning method (i.e. \textit{CGRL}) for inference of multi-network associations \cite{liu2016cross}. 
The Sylvester equation is widely adopted in solving multi-network mining tasks, such as network alignment \cite{zhang2018attributed}, \cite{chu2019cross}, cross-network similarity learning \cite{li2019graph}, and social recommendation \cite{tang2013social}. Chen et al. propose a Sylvester equation-based solver for semi-supervised multi-label learning problems \cite{chen2008semi}. Du et al. propose a Krylov-subspace based fast solver (i.e. \textit{FASTEN}) for the Sylvester equation for various multi-network mining tasks. 

\noindent \textbf{B - Graph Neural Network Models.}
Graph Neural Networks (GNN) include a broad range of deep learning models on network data. Recent advances primarily concentrate on convolutional models, such as \textit{GCN} \cite{kipf2016semi}, \textit{GAT} \cite{velivckovic2017graph}, \textit{GIN} \cite{xu2018powerful} and \textit{GraphSAGE} \cite{hamilton2017inductive}. We will briefly review representative GNN models related to multi-network mining tasks and matrix completion. Multi-Graph CNN (\textit{MGCNN}) by Monti et al. \cite{monti2017geometric} is one of earliest works which explores the convolutional models on multi-networks for matrix completion. \textit{GC-MC} by Berg et al. \cite{berg2017graph} proposes a network-based auto-encoder framework for matrix completion. The model produces latent features of users and items through message passing on the bipartite interaction networks. \textit{IGMC} by Zhang et al. \cite{zhang2019inductive} proposes an inductive matrix completion method using GNN model on bipartite graphs induced from user-item ratings when the side information is unavailable. 
\textit{GraphRec} by Fan et al. \cite{fan2019graph} jointly captures the interactions and opinions in the user-user and user-item network and propose a GNN-based framework for recommendation.


\vspace{-0.5\baselineskip}
\section{Conclusion}\label{sec:conclusion}
In this paper, we propose \symgnn, a flexible neural framework for generalizing the traditional Sylvester equation towards an end-to-end neural model for multi-network mining. We further propose two specific instantiations of the \symgnn\ framework for geometric matrix completion task. 
The experimental results show that the proposed models overall outperform baselines on all existing benchmark datasets. Furthermore, with slight performance drop, the proposed low-rank instantiation reduces the memory consumption by $16.98\%$ on average. 


\hide{
This material is supported by the National Science Foundation under Grant No. IIS-1651203, IIS-1715385, IIS-1743040, and CNS-1629888, by DTRA under the grant number HDTRA1-16-0017, by the United States Air Force and DARPA under contract number FA8750-17-C-0153\footnote{Distribution Statement "A" (Approved for Public Release, Distribution Unlimited)}, by Army Research Office under the contract number W911NF-16-1-0168, and by the U.S. Department of Homeland Security under Grant Award Number 2017-ST-061-QA0001. The content of the information in this document does not necessarily reflect the position or the policy of the Government, and no official endorsement should be inferred. The U.S. Government is authorized to reproduce and distribute reprints for Government purposes notwithstanding any copyright notation here on.
}

\bibliographystyle{style_files/named}
\bibliography{reference.bib}


\end{document}